\documentclass[letterpaper, 10pt]{criarticle}

\usepackage{amsmath} 
\usepackage{amsthm} 
\usepackage{amsfonts} 
\usepackage{amssymb}
\usepackage[inline]{enumitem}

\usepackage[numbers,sort&compress]{natbib}

\usepackage{titlesec}
\usepackage{verbatim}
\usepackage{booktabs}
\usepackage{multirow}
\usepackage{dcolumn}
\usepackage{array}
\usepackage{url}
\usepackage{calc}

\usepackage{authblk}
\usepackage{color}
\usepackage{graphicx}
\usepackage[verbose]{wrapfig}

\usepackage[margin=1in]{geometry}
\usepackage{caption} 
\DeclareCaptionType{copyrightbox}
\usepackage[T1]{fontenc}
\usepackage{txfonts}

\titlelabel{\thetitle.\quad}
\titleformat*{\subsection}{\normalsize \itshape}

\begin{document}
\date{\vspace{-3ex}}

\title{\begin{center}\Large \textbf{Towards Understanding the Generalization of Medical Text-to-SQL \\Models and Datasets} \end{center}}
\author{\textbf{Richard Tarbell M.S.$^1$, Kim-Kwang Raymond Choo, Ph.D.$^2$, Glenn Dietrich, Ph.D.$^2$, and}\\\textbf{Anthony Rios, Ph.D.$^2$}}
\affil{\textbf{$^1$Data Analytics, The University of Texas at San Antonio, USA}}
\affil{\textbf{$^2$Information Systems \& Cyber Security, The University of Texas at San Antonio, USA}}


\maketitle
\thispagestyle{empty}
\section*{Abstract}
\noindent \textit{Electronic medical records (EMRs) are stored in relational databases. It can be challenging to access the required information if the user is unfamiliar with the database schema or general database fundamentals. Hence, researchers have explored text-to-SQL generation methods that provide healthcare professionals direct access to EMR data without needing a database expert. However, currently available datasets have been essentially ``solved'' with state-of-the-art models achieving accuracy greater than or near  90\%. In this paper, we show that there is still a long way to go before solving text-to-SQL generation in the medical domain. To show this, we create new splits of the existing medical text-to-SQL dataset MIMICSQL that better measure the generalizability of the resulting models. We evaluate state-of-the-art language models on our new split showing substantial drops in performance with accuracy dropping from up to 92\% to 28\%, thus showing substantial room for improvement. Moreover, we introduce a novel data augmentation approach to improve the generalizability of the language models. Overall, this paper is the first step towards developing more robust text-to-SQL models in the medical domain.\footnote{The dataset and code will be released upon acceptance.}
}

\section*{Introduction}

Electronic medical records (EMRs) are crucial for evaluating and treating patients. For instance, EMRs can be used to predict mortality risk for patients~\cite{darabi2018forecasting,alves2018dynamic,tan2019ua} and is the basis of knowledge used for billing~\cite{rios2018emr} (e.g., with ICD10 codes). Hospitals generally store EMRs in relational databases. Structured query languages (e.g., SQL) are used to access items within the database. Hence, to access the data, users must know the database structure (e.g., the tables and columns) and the values stored in each column (e.g., understanding ICD10 codes). Suppose practitioners need to extract relevant knowledge from EMR-related databases. In that case, they need to work with database experts because it may be difficult for medical professionals to gather the information required efficiently, particularly as databases change. Furthermore, it is challenging for database experts to keep track of everything as it ages and changes over time (e.g., because of changes in diagnosis code standards), thus increasing the cost of maintenance. In this paper, we explore the text-to-SQL task, which can help facilitate easy access to medical information in databases by users who are not experts and make the information extraction easier for the database maintainers.

\textit{How can we reduce the expertise needed to query these databases? }Recent research has focused on developing a natural language interface for these databases that could significantly improve accessibility by allowing users to retrieve and use the information without programming expertise. Researchers have made substantial advances in recent years by releasing large-scale datasets and models for text-to-SQL generation~\cite{yu2018spider,wang2020text}. Developing a system that can accept a practitioner's question in a text-based format that automatically creates an accurate corresponding SQL query to return accurate answers allows professionals to get much-needed answers quickly and easily.

\citet{wang2020text} created the \textit{MIMICSQL} text-to-SQL dataset in a two-stage manner. First, they developed templates to create text—SQL tuples automatically (e.g., ``How many patients have language [BLANK]'', where [BLANK] is filled in with various languages mentioned in the database). Second, they paraphrased the text in the automatically generated templates to add more variety (i.e., so the text was not always specified in the same manner). \citet{wang2020text} evaluated their models on the original template data and the paraphrased version, achieving a logic form and execution accuracy of .912 and .940, respectively, on the template-based test set. In addition, they achieve a logic form and execution accuracy of .556 and .654, respectively, on the paraphrased data. This result shows that the paraphrased data provides a unique test bed beyond the template data to measure how well the models handle differences/variations in the questions. Nevertheless, \citet{pan2021} further improved on the work of \citet{wang2020text} by introducing a BERT-based method to generate better SQL statements resulting in scores of .784 and .899 on the paraphrased test set in terms of logic form and execution accuracy, respectively.

Current accuracy results on the MIMICSQL dataset are in the 90s for the template-based data and nearly .90 accuracy for their paraphrased dataset (i.e., a variation of the template-based question-SQL pairs where humans have rewritten the questions). Does this mean that medical text-to-SQL is (nearly) solved? Unfortunately, the current MIMICSQL train-test splits do not measure the model's generalizability toward understanding the structure of the database. Many of the training and test data questions are just variations, e.g., ``\textit{how many patients whose language is cape?}'' vs. ``\textit{how many patients whose language is port?}''. Hence, for a new question in the test set, the model only needs to understand which part of the text represents the conditional value, given it sees slight variations in the training data. What happens when new tables or columns are added? What if the values change for a specific column (e.g., transferring to the next version of ICD codes)? How would the models perform at hospitals where the database structure is likely substantially different? For other biomedical natural language processing (NLP) tasks, prior research in medical-related machine learning applications has shown substantial differences in the cross-hospital performance~\cite{rios2019cross,ding2021semi}. Similarly, this research focuses on changing and reevaluating new splits of the MIMICSQL that can help measure generalizability toward identifying better and more robust medical text-to-SQL models.

Another limitation of the MIMICSQL dataset is the lack of diversity in the input queries. While there have been efforts for paraphrasing (i.e., human rewording) the automatically generated (template-based) input instances, the diversity of the questions is limited by the imagination of the annotators. For instance, the query ``how many patients whose language is hait?'' is generally transformed into a phrase like ``find the number of patients who prefer haitian language,'' with the new version used often for similar questions (i.e., asking about different languages) in the dataset. Hence, models trained on the paraphrased dataset still struggle to generalize to new question variations. Therefore, we investigate the research question, can we generate models using the template data directly, ignoring the paraphrased data during training? This evaluation setting can better measure how well the models can perform for new question phrasings. Moreover, it is costly to paraphrase questions at scale for template-generated text-to-SQL datasets. If similar performance can be achieved without the additional cost, then integration with new database formats and EHR systems will be more feasible at scale. We explore two methods of improving generalizability. First, we investigate training on additional non-medical text-to-SQL datasets. We hypothesize that adding more diverse training data, even if it is not medical related, can improve the ability for the model to generalize over changes in database schemas. Second, we explore generating new synthetic data using back-translation. The phrasings are limited, particularly for the template-based MIMICSQL data. back-translation-based synthetic data has shown to generate useful synthetic data for other NLP tasks (e.g., correcting translation errors)~\cite{lee2021adaptation}. Thus, we explore its use for text-to-SQL generation in the medical domain.

In this paper, we seek to answer the following research questions:
\begin{enumerate}[label=\bfseries RQ \arabic*.~, leftmargin=*] 
    \item Are there issues with current medical text-to-SQL dataset train/dev/test partitions that limit the measure of model generalizability? We evaluate state-of-the-art language models on the original data splits to answer this question. Moreover, we  generate new train/dev/test splits for the MIMICSQL dataset called ``MIMICSQL 2.0’’. The dataset is a \textit{first step} towards finding and developing more robust and generalizable medical text-to-SQL models. 
    \item Can we improve the generalizability of state-of-the-art language models for medical text-to-SQL tasks using data augmentation? If novel synthetic data sources or additional data can match the training performance on manually paraphrased examples, this can reduce the cost of deploying these models to various EHR systems. Moreover, as new data is integrated, new templates and synthetic data can be introduced to add new capabilities to the model with minimal human annotation requirements.
\end{enumerate}



\section*{Related Literature}

This section describes the two main areas of relevant prior work, namely: Text-to-SQL Dataset Construction and Medical Text-to-SQL. 

\paragraph{Text-to-SQL Dataset Construction.}
There has been a recent surge toward better evaluation of text-to-SQL systems, particularly in measuring their generalizability. However, little has been explored in the medical domain.
Text-to-SQL is a semantic parsing task, i.e., generating a machine-understandable representation of the text’s meaning. Many datasets have been created that take an input (text) and generate outputs in many formats (e.g., SQL and logic forms). The present discourse provides a brief overview of various datasets, including the Airline Travel Information System (ATIS)\cite{data-atis-original,data-atis-geography-scholar}, Geography\cite{data-geography-original,data-atis-geography-scholar}, Restaurants~\cite{data-restaurants, data-restaurants-logic, data-restaurants-original}, WikiSQL~\cite{zhong2017seq2sql}, Spider~\cite{yu2018spider}, IMDB and Yelp~\cite{data-sql-imdb-yelp}. The ATIS dataset encompasses a collection of questions and their corresponding SQL queries about airline travel, encompassing flight schedules, ticket prices, and seat availability. The Geography dataset, on the other hand, comprises questions and SQL queries related to geographic locations, encompassing parameters such as population, area, and coordinates. The Restaurants dataset entails text and SQL queries about restaurant information, such as menu items, prices, and reviews. The Scholar dataset contains text and SQL queries related to academic research, including author names, publication titles, and citations. Finally, the IMDB and Yelp datasets encompass questions and SQL queries related to movie information, such as titles, release dates, ratings, and restaurant and business information, including ratings, reviews, and opening hours~\cite{data-sql-imdb-yelp}.


The WikiSQL~\cite{zhong2017seq2sql} dataset contains natural language questions and their corresponding SQL queries from the domain of Wikipedia tables, such as population data and historical events. Interestingly, because of the size of the Wikipedia dataset, databases/tables that appear in the training data never appear in the test or development datasets. Thus, models developed on WikiSQL must be able to take a table schema and question as input, then return the relevant query, thus measuring the generalizability of the models. This is important given the size of Wikipedia and the rapid nature in which it changes. However, to generate many text-SQL tuples for training/evaluation, \citet{zhong2017seq2sql} made simplified assumptions about the SQL queries and databases. The WikiSQL dataset’s SQL components only cover one \texttt{SELECT} column at a time and aggregation with \texttt{WHERE} conditions. Moreover, all the databases only contain single tables, there is no \texttt{JOIN}, \texttt{GROUP BY}, and \texttt{ORDER BY}, or other complex operations. \citet{yu2018spider} expands on the work by \citet{zhong2017seq2sql} to create a new text-to-SQL dataset (Spider) that contains more complicated queries. The SQL queries in the Spider dataset contain nested queries and clauses like \texttt{GROUP BY} and \texttt{HAVING}, which are far more complicated than that in another well-studied cross-domain benchmark, WikiSQL~\cite{zhong2017seq2sql}. This work explores the Spider dataset as an additional training corpus.

The major text-to-SQL dataset within the healthcare domain {MIMICSQL}~\cite{wang2020text} was created with the widely used Medical Information Mart for Intensive Care III (MIMIC III) corpus~\cite{johnson2016mimic}. The dataset was generated in a two-stage fashion. First, text-SQL pairs were automatically generated using templates. Second, freelancers reworded 10,000 questions to create a simultaneous dataset of Natural Language (NL) questions where they paraphrased the template-produced text.


\paragraph{Text-to-SQL Methods.}
Text-to-SQL is an increasingly popular research area in natural language processing that focuses on automatically translating natural language queries into structured SQL queries. Recent advances in neural network-based models have enabled Text-to-SQL systems to achieve ever higher levels of accuracy and performance. 
Many researchers have introduced novel text-to-SQL models or evaluated the limits of existing models~\cite{xu2017sqlnet,zhong2017seq2sql,bogin2019representing,wang2018robust,yu2018typesql,scholak2021picard,xie2022unifiedskg,wang2022proton,chen2021evaluating,sun2022leveraging}. \citet{wang2020rat} proposed a unified framework, based on relation-aware self-attention, to address schema encoding, schema linking, and feature representation, and achieve state-of-the-art performance on the challenging Spider dataset, with an exact match accuracy of 65.6\% when augmented with BERT. \citet{scholak2021picard} introduced PICARD, a method to constrain language models during fine-tuning on formal languages like SQL. It uses incremental parsing to reject invalid tokens and find valid output sequences. PICARD improves T5 models' performance on text-to-SQL translation tasks, making them state-of-the-art solutions on the challenging Spider and CoSQL datasets. multi-task knowledge grounding~\cite{xie2022unifiedskg}. \citet{wang2022proton} proposed a novel framework for schema linking in text-to-SQL parsers that utilizes a probing procedure based on Poincaré distance metric to elicit relational structures from pre-trained language models. The results show that the framework outperforms rule-based methods for schema linking and sets a new state-of-the-art performance on three benchmarks, indicating that the probing procedure can robustly capture semantic correspondences. \citet{rajkumar2022evaluating} evaluated the Codex language model's~\cite{chen2021evaluating} text-to-SQL capabilities on several benchmarks without any finetuning and finds that it performs competitively with an execution accuracy of up to 67\% on the Spider benchmark. The authors also demonstrated that providing a small number of in-domain examples in the prompt can enable Codex to outperform state-of-the-art models that are finetuned on such few-shot examples.

With respect to medical text-to-SQL models, \citet{wang2020text} also introduced a novel method for medical text-to-SQL called TREQS. They evaluated TREQS for both execution accuracy and logical accuracy. TREQS achieved an accuracy greater than .91 for both logical and execution on their test split and .85 logical and .92 execution accuracies on the development split. In this work, we revisit the MIMICSQL dataset to explore issues with generalization and propose a new split to improve robust medical text-to-SQL model design and evaluation. Expanding on the work by \citet{wang2020text}, \cite{pan2021} presented a novel approach to medical text-to-SQL generation. The proposed model, named MedTS, uses a pretrained BERT model as the encoder and a LSTM-based decoder. MedTS benefits from the multi-head attention mechanism of the pretrained encoder, enabling it to capture the semantic and dependency relationships between the textual question and the database schema. The grammar-based decoding strategy of MedTS reduces the search space and generates a tree-structured intermediate representation by incorporating predefined grammatical rules. Compared to TREQS, the proposed model outperforms in terms of logic form and execution accuracy, achieving scores of .784 and .899, respectively, on the paraphrased MIMICSQL test set. 

\begin{figure*}
    \centering
    \includegraphics[width=.8\linewidth]{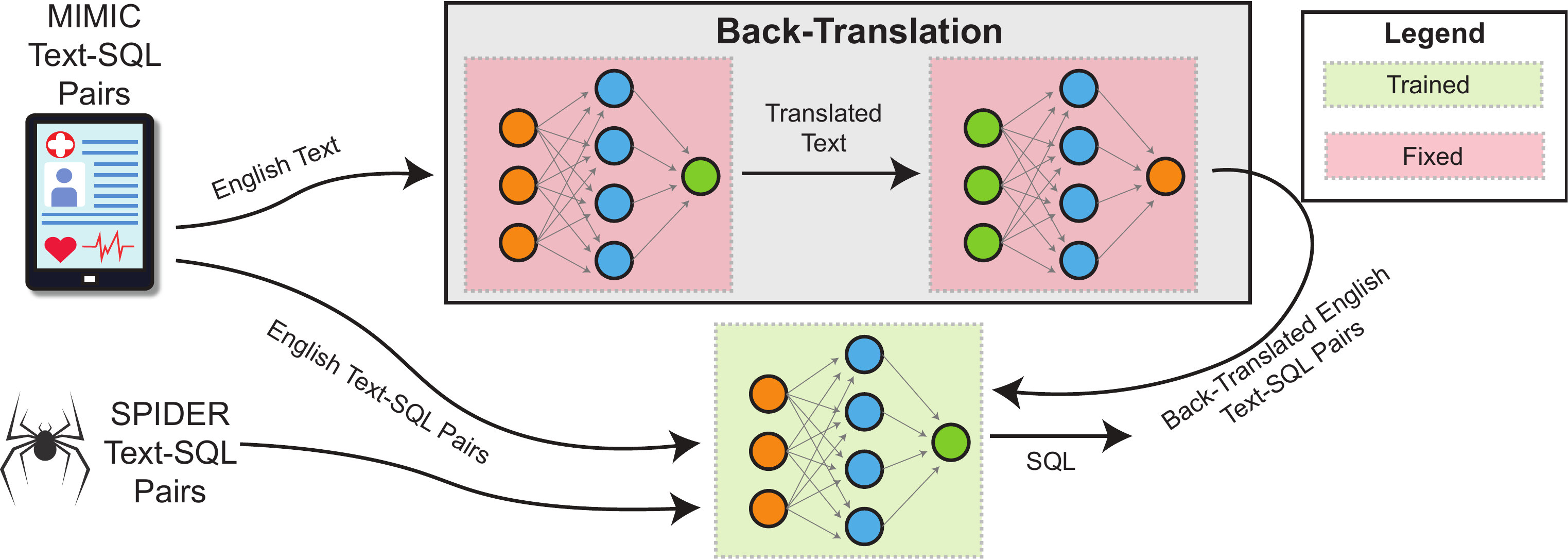}
    \caption{Model overview.}\vspace{-1.5em}
    \label{fig:overview}
\end{figure*}

The most similar work to ours is by \citet{zhao-etal-2022-bridging}. They showed that existing (general domain) benchmarks, such as Spider and WikiSQL cannot capture specific out-of-domain generalization issues that are important for practical applications. This is like our study in which we argue that existing medical dataset splits cannot understand the schema of the medical databases and genuinely understand how to map the text back to the schema. Hence, our study aims to take similar ideas from \cite{zhao-etal-2022-bridging} to study gaps in measuring the generalizability of text-to-SQL models in the medical domain. Likewise, our approach to improve generalization follows similar ideas from \citet{wolfson2022weakly}. \citet{wolfson2022weakly} proposed a weak supervision approach to train text-to-SQL parsers without relying on annotated natural language and SQL pairs. Instead, the approach uses question-meaning representation called QDMR, and experimental results show that the weakly supervised models perform competitively with those trained on annotated text-SQL paired data. In our work, we explore the generation of automatic paraphrasing (synthetic data) via the use of back-translation~\cite{sennrich2016improving}.


\section*{Datasets and Methods}

This section summarizes the datasets and methods used in this paper. We provide an overview of our multifaceted method in Figure~\ref{fig:overview}, which comprises three main components. First, the main component of our system is the T5 language model, which takes a database schema and question as input and generates a SQL statement. Second, we introduce a data augmentation method using back-translation to add more diversity to the text questions. Third, we train both on the MIMICSQL and Spider datasets.

\subsection*{Datasets}
We evaluate using the original MIMICSQL dataset and our new partition MIMICSQL 2.0. We also describe the additional training data, Spider, we used to complement the MIMICSQL 2.0 dataset to improve model performance.

\vspace{2mm} \noindent \textbf{MIMICSQL.} MIMICSQL is based on the MIMIC-III (Medical Information Mart for Intensive Care) dataset, a collection of electronic medical records from over 40,000 patients collected between 2001 and 2012 from the critical care units of the Beth Israel Deaconess Medical Center. The data includes information on patient demographics, vital signs, lab results, medications, clinical notes, and ICD-9 diagnostic and procedure codes. This dataset has been widely used in research on various critical care topics such as ICD-9 coding~\cite{rios2018emr}, mortality risk prediction~\cite{darabi2018forecasting,alves2018dynamic,tan2019ua}, acute respiratory distress syndrome, and patient outcomes prediction. Due to its large size and breadth of information, MIMIC-III is considered a valuable resource for researchers studying critical care and health informatics. MIMIC-SQL is a version of the MIMIC-III dataset (i.e., containing a subset of MIMIC-III) that is structured in a relational database format. The database statistics for the entire dataset from MIMICSQL are shown in Table~\ref{table:ta2}. The size of the training, development, and test splits in the original dataset are shown in Table~\ref{tab:splits}.

\begin{wraptable}[12]{r}{.5\linewidth}
\vspace{-1.5em}
\centering
\resizebox{\linewidth}{!}{%
\begin{tabular}{ll}
    \toprule
    \textbf{Type} & \textbf{Stats} \\
    \midrule
    \textbf{Patients} & 46,520\\
    \textbf{Tables} & 5\\
    \textbf{Columns in tables} & 23/5/5/7/9\\
    \textbf{Question-SQL Pairs} & 10,000\\
    \textbf{Avg. Template Question Length (in words)} & 18.39\\
    \textbf{Avg. Rephrased Question Length (in words)} & 16.45\\
    \textbf{Avg. SQL Query Length }& 21.14\\
    \textbf{Avg. Aggregation Columns} & 1.10\\
    \textbf{Avg. conditions} & 1.76\\ \bottomrule
\end{tabular}}
 \caption{MIMICSQL Database Statistics.}
\label{table:ta2}
\end{wraptable}

\vspace{2mm} \noindent \textbf{New MIMICSQL 2.0.} Our goal is two-fold. First, we want to generate new splits that better measure the generalizability of the medical text-to-SQL models. More importantly, the goal is to create a more challenging split that is not as easily solvable—the idea for our new split forces the model to reason about tables. Second, we want the training splits to be roughly the same size as the original data to ensure enough data for training and evaluation. Based on ideas from the WikiSQL dataset~\cite{zhong2017seq2sql}, we look to partition the dataset based on the tables to measure generalizability. However, our rule differs in one important way. The WikiSQL dataset contains 24241 tables, whereas MIMICSQL only contains five tables. Hence, ensuring that a table only appears in one split is impossible, particularly if we want to ensure the new dataset splits stay close to the original split sizes. Therefore, we choose three tables for the basis of our dataset modifications: PROCEDURES, PRESCRIPTIONS, and LAB.

\begin{wraptable}[7]{r}{.5\linewidth}
\vspace{-1.5em}
\centering
\resizebox{\linewidth}{!}{%
\begin{tabular}{@{}lllllll@{}}
\toprule
                   & \multicolumn{3}{c}{\textbf{Original}} & \multicolumn{3}{c}{\textbf{New Split}} \\ \cmidrule(lr){2-4} \cmidrule(lr){5-7}
                   & \textbf{Train}     & \textbf{Dev}    & \textbf{Test}    & \textbf{Train}     & \textbf{Dev}     & \textbf{Test}    \\ \midrule
\textbf{\# of Examples} &      8,000     &    1,000    &       1,000  &        8,346   &   796      &   1,000      \\ \bottomrule
\end{tabular}%
}
\caption{Number of examples in each data split into the original data vs. the new splits.}
\label{tab:splits}
\end{wraptable}

We remove each table from the training split in one of two ways. We remove it such that it does not appear as the main table in the query (i.e., as the argument to the \texttt{FROM} field), or we remove it such that it does not appear as part of an \texttt{INNER JOIN}, which makes up a large percentage of the queries in MIMICSQL. Specifically, if a query in the training data has the table ``PROCEDURES’’ in the following position within the query ``\texttt{INNER JOIN} PROCEDURES'' then our test/development sets do not have PROCEDURES following \texttt{INNER JOIN}, but only include following the \texttt{FROM} statement. Similarly, we moved all the statements with ``\texttt{FROM} LAB'' and ``\texttt{FROM} PRESCRIPTIONS’’ into the test/development sets while leaving all the ``\texttt{INNER JOIN} LAB'' and ``\texttt{INNER JOIN} PRESCRIPTIONS'' in the training set. Intuitively, if a model learns to understand the schema of the database and learns to map questions to the schema, then having a table appear in a \texttt{FROM} statement in the training set should provide the model with enough information to infer how to use it with \texttt{INNER JOIN} (and vise-versa in the other direction). Removing tables in both cases resulted in unbalanced training/development/test splits (e.g., with very few or too many examples in the test/development splits). Hence, this strategy balances, creating a split that measures generalizability with training/development/test set sizes.  The size of the training, development, and test splits in the new dataset are shown in Table~\ref{tab:splits}.

\vspace{2mm} \noindent \textbf{SPIDER.} The Spider text-to-SQL dataset is a collection of natural language questions and their corresponding SQL queries. It was created to facilitate research in text-to-SQL, which involves developing natural language interfaces to databases. The dataset consists of over 10,000 questions based on 200 complex databases from various domains, including geography, music, and film. Each question is annotated with its corresponding SQL query, which extracts information from the corresponding database. The dataset includes simple and complex queries and various types of SQL clauses and operators. The Spider dataset is widely used in developing and evaluating natural language interfaces to databases and is considered one of the benchmark datasets in the field. Hence, we use this as an auxiliary training dataset to explore generalization to the medical domain. We use the entire Spider dataset for training.

\subsection*{Methods}

\paragraph{T5 for Text-to-SQL}
The T5 (Text-to-Text Transfer Transformer) model is a large-scale pre-trained language model developed by Google, based on the Transformer architecture. With up to 11 billion parameters~\footnote{In our experiments, we use flan-t5-base from the HuggingFace package~\cite{wolf2019huggingface}.}, T5 is one of the largest language models currently available that runs on commodity hardware with publicly available weights. Unlike previous models trained for specific tasks, T5 is a text-to-text model that can be fine-tuned for various NLP tasks, including text classification, machine translation, question-answering, and summarization. To train T5, Google used a diverse corpus of text data from sources such as Wikipedia, Common Crawl, and the web pages crawled by Google.

We apply the T5 model to the medical text-to-SQL task. The input to the model includes the database schema and the question. The schema contains information about the tables and the columns within each table. Formally, let $t_i$ represent a table $i$, and let $c_{i,j}$ represent a column $j$ in table $i$. Each column has an attribute $a_{i,j}$ that represents the $j$-th column's datatype in table $i$. For instance, in MIMICSQL, we have the table ``DEMOGRAPHIC'', which stores information about patients (e.g., name, age, gender, language). Two columns within ``DEMOGRAPHIC'' include ``NAME'' and ``AGE''. The attribute assigned to the NAME column is ``text'' since it contains strings like ``John Doe''. The attribute assigned to ``AGE'' is ``number''. Given all of the tables, columns, and attributes in a database, we generate the schema represents in the form of $s = [*, t_1, c_{1,1}, a_{1,1}, c_{1,2}, a_{1,2}, t_2, c_{2,1}, a_{2,1}, c_{2,2}, a_{2,2}, \ldots]$. $*$ is used as a special symbol to represent the ``all columns''. In practice, this would look like ``* DEMOGRAPHIC NAME text AGE number DIAGNOSIS ICD\_CODE text'', where DIAGNOSIS is another MIMIC table and ICD\_CODE is a column within the DIAGNOSIS table. Given the schema $s$, we append the question $q$ as $[s, [SEP], q]$, where $q$ is a sequence of tokens forming the question (e.g., ``What is the age of John Doe?'') and $[SEP]$ is a special token that separates the schema from the query.

At inference time, we use execution-guided decoding, which has been shown to outperform standard beam-search-based decoding~\cite{wang2018robust}. Intuitively, the decoder generates SQL queries by taking into account the execution of the query on a database.  In our implementation, we use beam search with a beam size of ten to generate the ten queries with the highest joint probabilities. Next, we execute each of the queries best on the most probable to least probable and return the first query that executes correctly on the database. If all queries fail to execute, we return the first query, which will be counted as incorrect in our evaluation metrics.

\paragraph{Back-Translation for Data Augmentation}
Back-translation is a technique used in NLP for machine translation~\cite{sennrich2016improving}. It involves translating a text from one language into a second language, then translating the second language translation back into the original language. But, it can generalize to any task that involves translating from some source $X$ to target $Y$, then back to $X$~\cite{fan2022back}. In this work, we translate every question in the MIMICSQL corpus to French and German, then back to English. The back-translated English text is used as synthetic paraphrases for training, similar to the manually curated paraphrases used in the original MIMICSQL dataset manually. However, the paraphrases are automatically generated to add more diversity to the questions. For translation, we use the state-of-the-art models from Facebook~\cite{koishekenov2022memory}.

\paragraph{Out-of-Domain Training Data}
Multi-domain learning is a machine learning technique that involves training a single model to perform well across multiple related domains. In traditional machine learning approaches, a separate model would need to be trained for each domain, which can be time-consuming and resource-intensive and may miss out on common characteristics shared between the two domains. Specifically, training on multiple domains can potentially improve performance on other domains if the target domains (i.e., MIMICSQL) are limited with regard to the number of training examples. Hence, we explore jointly training on the Spider and MIMICSQL datasets to understand if out-of-domain data can improve medical text-to-SQL model performance.

\section*{Experiments}

\begin{table*}[t]
\centering
\resizebox{.85\linewidth}{!}{%
\begin{tabular}{lllllllll}
\toprule
& \multicolumn{4}{c}{\textbf{T} $\rightarrow$ \textbf{T}} & \multicolumn{4}{c}{\textbf{P} $\rightarrow$ \textbf{P}} \\ \cmidrule(lr){2-5} \cmidrule(lr){6-9}
                  & \multicolumn{2}{c}{\textbf{Dev}} & \multicolumn{2}{c}{\textbf{Test}}             &     \multicolumn{2}{c}{\textbf{Dev}} & \multicolumn{2}{c}{\textbf{Test}}                  \\ \cmidrule(lr){2-3} \cmidrule(lr){4-5} \cmidrule(lr){6-7} \cmidrule(lr){8-9}
                  & $ACC_{LF}$ & $ACC_{EX}$ & $ACC_{LF}$ & $ACC_{EX}$ & $ACC_{LF}$ & $ACC_{EX}$ & $ACC_{LF}$ & $ACC_{EX}$ \\ \midrule
\textbf{Coarse2Fine}~\cite{wang2020text} &  .298 &  .321 &  .518  & .526          & .217 &  .309 &  .378 & .496    \\
\textbf{M-SQLNET}~\cite{wang2020text}  & .258  & .588  & .382 & .603 & .086 & .225 &  .142 &  .260      \\
\textbf{Seq2Seq}~\cite{wang2020text}           & .098             & .372               & .160             & .323      & .076 & .112  & .091 & .131             \\
\textbf{Seq2Seq + recover}~\cite{wang2020text} & .138             & .429               & .231             & .397   & .092 & .195 &  .103 &  .173          \\
\textbf{PtrGen}~\cite{wang2020text}            & .312             & .536               & .372             & .506     & .126 & .174 & .160 & .222               \\
\textbf{PtrGen + recover}~\cite{wang2020text}  & .442             & .645               & .426             & .554    &  .181 & .325 & .180  & .292         \\
\textbf{TREQS}~\cite{wang2020text}           & .711             & .803               & .802             & .802   & .451 & .511 & .486 & .556           \\
\textbf{TREQS + recover}~\cite{wang2020text}   & .853             & .924               & .912             & \textbf{.940}   & .562 &  .675 & .556  & .654              \\ 
\textbf{MedTS}~\cite{pan2021}  & --- & --- & --- & --- & 	.681 &  .880 & .784 & .899\\ \midrule
\textbf{T5}              & \textbf{.932}             & \textbf{.937}                & \textbf{.916 }            & .923     &  .866            &       \textbf{.916}      &      \textbf{.899}     &         \textbf{.936}      \\  \bottomrule
\end{tabular}%
}
\caption{Logical form ($ACC_{LF}$) and execution accuracy ($ACC_{EX}$) on the MIMICSQL data splits.}\vspace{-1.5em}
\label{tab:orignal-main}
\end{table*}

In this section, we describe the evaluation metrics used in our study and the results/findings of our experiments.

\subsection*{Evaluation Metrics}
Following the work of \citet{wang2020text}, we use two major evaluation metrics: Logic Form Accuracy~\cite{zhong2017seq2sql} ($ACC_{LF}$) and Execution Accuracy ($ACC_{EX}$). $ACC_{LF}$ is calculated by comparing the generated SQL queries with the true SQL queries token-by-token. However, this is a strict way of measuring performance, e.g., if the order of boolean expressions is different, but the logic is the same, then $ACC_{LF}$ may overly penalize a model.

$ACC_{EX}$ is calculated based on the rows returned by both the generated and true queries, row-by-row. ``Execution accuracy'' measures the percentage of generated SQL queries that can retrieve the correct results from a database when executed. To evaluate the execution accuracy of a text-to-SQL generation model, a test dataset consisting of natural language text and corresponding SQL queries is used. The model is then used to generate SQL queries from the natural language text. These generated queries are executed against a database and compared to the expected results from the corresponding SQL queries. The execution accuracy is calculated as the percentage of generated queries that return the same results as the corresponding queries. For example, if a model generates 100 SQL queries, 96 of them return the same results as the corresponding queries, then the execution accuracy would be 96\%.

Execution accuracy is a good metric to evaluate the quality of the generated SQL statements, and it's also important to consider other metrics, such as the syntax and semantics of the generated SQL statements and the ability of the model to generalize to new test cases. This is a more lenient way of measuring the model performance since the model can generate incorrect SQL queries but still get the correct answer by luck. Hence, combining both is important for general text-to-SQL model evaluation.


\subsection*{Baseline Models}

\vspace{2mm} \noindent \textbf{TREQS.~\cite{wang2020text}}
TREQS is a deep learning approach that utilizes the popular sequence-to-sequence model to directly translate natural language questions into SQL queries. The model incorporates an attentive-copying mechanism and task-specific look-up tables to make necessary modifications to the generated query.

\vspace{2mm} \noindent \textbf{TREQS + recovery.~\cite{wang2020text}} The technique employs the use of the ROUGE-L string matching metric~\cite{lin-2004-rouge}, which calculates both word-level and character-level similarities between two sequences. The ROUGE-L metric is used to identify the most similar condition value from a look-up table for each predicted condition value. The predicted condition value is then replaced with the most similar value from the look-up table to obtain the exact condition value, thus enhancing the accuracy and executability of the generated SQL query.

\vspace{2mm} \noindent \textbf{Other Baselines.} For the original dataset, the results of the additional models reported by \citet{wang2020text} (e.g., Coarse2Fine and Seq2Seq) for a comprehensive listing. Moreover, we compare to MedTS introduced by \citet{pan2021}.

\vspace{2mm} \noindent \textbf{Our Baselines}. We experiment with training only on the MIMICSQL datasets (T5), training using back-translation (T5-back-translation), training on both MIMICSQL and Spider (T5 + Spider), and a combination combining all approaches (T5 + Spider + back-translation). 

\subsection*{Results}

\noindent \textbf{RQ 1. Are there issues with the current medical text-to-SQL dataset's train/dev/test partitions that limit model generalizability?} In Table~\ref{tab:orignal-main}, we report the results of the prior state-of-the-art on the original MIMICSQL data split for the template-based (Models trained and tested on template-generated data) and paraphrased-based data (i.e., models trained and tested on paraphrased data). First, for the Template-to-Template (T $\rightarrow$ T) data, we find that both the previous work (TREQS) and base T5 model is able to achieve logical and execution accuracies over .9. The results are high for both the development and test datasets. This result indicates that both modern models (T5) and older methods generally ``solve'' the datasets. Meaning, that the model can extract the conditional values from the text and generate the relevant SQL statement with the extracted value based on what was seen in the training dataset.

We also report the training and evaluation results on paraphrased data in Table~\ref{tab:orignal-main} (P $\rightarrow$ P). Many of the prior results were much lower than the template dataset. For example, TREQS + recover received an $ACC_{EX}$ of .940 on the template-based test set. Yet, it received an F1 of only .654 on the paraphrased dataset. This result indicates the paraphrased data, which is much more diverse than the templates, is more difficult for the models. However, recent work by MedTS improves on TREQS + recover substantially, achieving an $ACC_{LF}$ of .784 and an $ACC_{EX}$ of .899. Moreover, our model T5 improves on the results even further with a test result of .899 $ACC_{LF}$ and .936 $ACC_{EX}$. This indicates, again, that the original dataset. Yet, even though there are more diverse questions in the paraphrased dataset, there are examples where the underlying meaning of the questions (with very similar SQL statements) appears within both the text and training datasets. Hence, models can learn these patterns to ``solve'' the dataset.

\begin{table*}[t]
\centering
\resizebox{.75\linewidth}{!}{%
\begin{tabular}{lllllll}
\toprule
& \multicolumn{2}{c}{\textbf{T} $\rightarrow$ \textbf{T}} & \multicolumn{2}{c}{\textbf{P} $\rightarrow$ \textbf{P}}  & \multicolumn{2}{c}{\textbf{T} $\rightarrow$ \textbf{P}} \\ \cmidrule(lr){2-3} \cmidrule(lr){4-5}  \cmidrule(lr){6-7}
                  & $ACC_{LF}$ & $ACC_{EX}$ & $ACC_{LF}$ & $ACC_{EX}$  & $ACC_{LF}$ & $ACC_{EX}$  \\ \midrule
\textbf{TREQS}         & .063            & .166      &  .068 & .173  & .052       & .154 \\
\textbf{TREQS + recover} &     .086            & .164     & .071 & .184  & .061 & .159   \\ \midrule
\textbf{T5}            & .149 &      .278  & .130  & .503 & .085 & .259 \\
\textbf{T5      + back-translation}         & .131 & .511 & .144  & .519  & \textbf{.208 }& .487 \\
\textbf{T5      + Spider}            & \textbf{.777} &  \textbf{.839}    & .134 &  .489 & .187  & .399 \\ 
\textbf{T5      + Spider + back-translation}                     & .196 &          .482 & \textbf{.233} & \textbf{.528} & .139   & \textbf{.501}    \\  \bottomrule
\end{tabular}%
}
\caption{Logical form ($ACC_{LF}$) and execution accuracy ($ACC_{EX}$) on the new MIMICSQL 2.0 data splits}
\label{tab:new-main}\vspace{-1.5em}
\end{table*}

\vspace{2mm} \noindent \textbf{RQ 2. Using data augmentation, can we improve the generalizability of state-of-the-art language models for medical text-to-SQL?} In Table~\ref{tab:new-main}, we report the overall results for our new  data split, MIMICSQL 2.0. Moreover, we report of training and evaluating on the template data (T $\rightarrow$ T), training and evaluating on the paraphrased data (P $\rightarrow$ P), and training on the template data and evaluating on the paraphrased data (T $\rightarrow$ P). First, for the template-to-template results (T $\rightarrow$ T), we find that  the overall results are much lower when compared to the original data split. Using the base T5 model, we only achieve an $ACC_{LF}$ of .149 and an $ACC_{EX}$ of .278 for the test dataset. Interestingly, we obtain high performance for the T5 + Spider model, achieving an $ACC_{LF}$ of .777 and an $ACC_{EX}$ of .839. We hypothesize that some of the templates are phrased in such a way that is similar enough you Spider-specific phrasing that the model is able to perform very well. Also, note that $ACC_{LF}$ and $ACC_{EX}$ are not entirely correlated. This is because the logical form accuracy $ACC_{LF}$ will mark instances as incorrect when columns, tables, or joins are in a different order, even if the logic is correct (e.g., ``SELECT A,B from TABLE'' and ``SELECT B,A from TABLE'' would result in an incorrect prediction using logical accuracy). Likewise, $ACC_{EX}$ may mark a query as correct if the returned results match the ground-truth, even if the underlying logic of the query is not correct. Yet, we can still see general patterns in the data. Using back-translation and the additional Spider dataset improves the results.

We find similar patterns for the P $\rightarrow$ P results. Again, performance is much worse than when we trained and evaluated the  models using the original training splits.  Moreover, we make two major findings. First, the Spider dataset does not result in accuracies as high as the T $\rightarrow$ T results. Again, we hypothesize that there are random characteristics of the question in the template dataset that the Spider training approach picked up on. Second, the general performance measures are relatively similar across the board. We obtained slight improvements with back-translation and back-translation+Spider, but the original model performs okay for the paraphrased data, while it underperformed substantially for the template data. One reason is that training on the template data alone results in overfitting issues at test time. Upon analyzing many of the errors, we found that the template model would make joins between tables that were incorrect by matching only joins that appear within the training data.

Finally, for the T $\rightarrow$ P results, we find that the base T5 model, when trained on the template data alone, performs poorly on the paraphrased data (.503  vs. .259 $ACC_{EX}$). However, when integrating data augmentation methods, we obtain accuracies that nearly match the results found when training on the paraphrased dataset directly (e.g., .519 vs. .487 $ACC_{EX}$ for T5 + back-translation). This finding indicates that if we need to generalize to new tables or database schemas, we can potentially generate data automatically using templates, add additional auxiliary training datasets and use back-translation to achieve adequate performance. This has the potential to reduce training costs and make integrating into various EHR systems more feasible.


\section*{Conclusion}

In this paper, we created a new training/development/test split of the MIMICSQL dataset. In the new split, all methods perform worse than the original split, thus providing a strong test-bed for future innovation.  Furthermore, we evaluated data augmentation methods by integrating external datasets (Spider) and via back-translation, improving model performance in general. We also want to point to two major avenues of future work. First, more work is needed to create a dataset with more complex queries. While the MIMICSQL dataset contains JOINS, it does not contain nested queries or highly complex queries as defined in the Spider dataset~\cite{yu2018spider}. Second, more work is needed to measure cross-hospital performance, particularly on real-world hospital databases. While MIMICSQL provides an excellent test bed because of its public availability, it does not match databases seen in practice~\cite{rios2019cross}. Hence, measuring and understanding cross-hospital performance is essential for broad applicability in the medical domain.

\section*{Acknowledgment}
Project sponsored by the National Security Agency under Grant / Cooperative Agreement (NCAE-C Grant) Number H93230-21-1-0172. The United States Government is authorized to reproduce and distribute reprints notwithstanding any copyright notion herin.

%
\bibliographystyle{van-auth-year}
\bibliography{scoonerdb}  

`\end{document}